\documentclass[9pt]{article}
\usepackage{epsfig}
\usepackage{epstopdf}
\usepackage{spconf,amsmath,amsthm,graphicx,amssymb, algorithm, algorithmic}
\usepackage{fancyhdr, wrapfig,caption,color}

\usepackage{MnSymbol}

\newcommand{\ben}{\begin{eqnarray}}

\newcommand{\een}{\end{eqnarray}}

\newcommand{\transpose}{^{\top}}
\newtheorem{thm}{Theorem}

\newcommand{\tr}{\mbox{tr}}

\newcommand{\la}{\langle}
\newcommand{\ra}{\rangle}

\newcommand{\cA}{\mathcal{A}}
\newcommand{\Poi}{\mbox{Poisson}}
\newcommand{\myvec}{\mbox{vec}}
\newcommand{\diag}{\mbox{diag}}

\title{Fast Algorithm for Low-Rank Matrix Recovery in Poisson Noise}
\name{Yang Cao and Yao Xie\thanks{Submitted to IEEE GLOBALSIP 2014, May 2014.} }
\address{Industrial and System Engineering, Georgia Institute of Technology}
\begin{document}
%
\maketitle
\begin{abstract}
This paper describes a new algorithm for recovering low-rank matrices from their linear measurements contaminated with Poisson noise: the Poisson noise Maximum Likelihood Singular Value thresholding (PMLSV) algorithm.
We propose a convex optimization formulation with a cost function consisting of the sum of a likelihood function and a regularization function which is proportional to the nuclear norm of the matrix. Instead of solving the optimization problem directly by semi-definite program (SDP), we derive an iterative singular value thresholding algorithm by expanding the likelihood function. We demonstrate the good performance of the proposed algorithm on recovery of solar flare images with Poisson noise: the algorithm is more efficient than solving SDP using the interior-point algorithm and it generates a good approximate solution compared to that solved from SDP.
\end{abstract}

\section{Introduction}

Recovery of a matrix $M$ from its linear measurements (or linear projections) contaminated with Poisson noise arises from various important applications such as optical imaging, nuclear medicine and X-ray imaging \cite{brady2009optical}. When $M$ is low-rank, we can still recover $M$ from a relatively small number of measurements, and it has been shown that under certain conditions $M$ can be recovered exactly \cite{candes2009exact}.

While there has been much success for low-rank matrix recovery and completion without noise or with additive Gaussian noise, relatively fewer results are available when the measurements are contaminated with Poisson noise \cite{xielow, Similar2014}. In \cite{xielow}, the authors establish some performance bounds without developing algorithms.  In \cite{Similar2014}, the parameters of Poisson distribution is sparse coefficients under certain compression dictionary, which is different with that of our model. The problem with Poisson noise is different because unlike Gaussian noise which has static noise variance, the variance of Poisson noise is proportional to the signal intensity. Also, we need to use a non-linear likelihood function to replace the $\ell_2$ norm penalty for data fitting term in the formulation.
Moreover, in practical systems with Poisson noise, many physical constraints have to be taken into consideration when recover the signal, e.g. the positivity of the signal and the total intensity constraint.

In this paper, we present a regularized maximum likelihood estimator to recover a low-rank matrix from linear measurements contaminated with Poisson noise. Instead of directly solving the convex optimization problem formulated this way, we present a generalized iterative singular value thresholding method \cite{cai2010singular}, which can be viewed as a consequence of approximating the log likelihood function by its second order Taylor expansion. The good performance of the proposed algorithm is demonstrated via numerical examples where we recover solar flare images with low-rank structure from Poisson measurements. We show that the proposed method is more efficient than solving the convex optimization using interior point method and it has good accuracy.

\section{Formulation}

\subsection{Model}
Suppose we wish to recover a matrix $M^* \in \mathbb{R}_+^{m_1\times m_2}$ consisting of nonnegative entries from $N$ linear measurements with Poisson measurements $y_i\in \mathbb{Z}_+^N$ that take the forms of
\ben
y_i \sim \Poi ([\mathcal{A} M^*]_i), \quad i = 1, \ldots, N,
\een
where the linear operator $\mathcal{A}: \mathbb{R}_+^{m_1\times m_2} \rightarrow \mathbb{R}^{N}$ models the measuring process of physical devices and it takes the following form:
\ben
[\mathcal{A} M]_i = \la A_i, M \ra \triangleq \tr(A_i\transpose M),
\een
where $A_i \in \mathbb{R}^{m_1\times m_2}$, and $\tr(X)$ denotes trace of a matrix $X$. In optical systems, the matrix $A_i$ models the masks that are applied to the light field before the intensity is measured. Let  $\myvec(X) = [x_1\transpose, \ldots, x_n\transpose]\transpose$ denote vectorized version of the matrix $X = [x_1, \ldots, x_n]$. Note that if we define
\[
A \triangleq \begin{bmatrix}
\mbox{vec}(A_1)\transpose \\
\vdots\\
\mbox{vec}(A_N)\transpose
\end{bmatrix}, \quad f \triangleq \myvec(M),
\]
then the measurements can be written as
\[
\mathcal{A} M = A f.
\]

We make the following assumptions about the system. Let $[X]_{ij}$ denote the element of matrix $X$ in the $i$th row and $j$th column, and  $[x]_j$ denotes the $j$th element of a vector $x$. Define the norm $\|X\|_{1, 1} = \sum_{i}\sum_j [X]_{ij}.$
By this notation, we assume that the total intensity of $M^*$, given by
\ben
I \triangleq \|M^*\|_{1, 1}
\een
is known a priori.
Also, to have physically realizable linear optical systems, we assume that the measurement operator $\mathcal{A}$ satisfies the following constraints:
\begin{enumerate}
\item (positivity-preserving) $[M]_{ij} \geq 0$ for all $i$, $j$ $\Rightarrow$ $[\mathcal{A} M]_i \geq 0$, for all $i$,
\item (flux-preserving) $\sum_{i=1}^N [\mathcal{A} M]_i \leq \|M\|_{1, 1}$.
\end{enumerate}
Our goal is to estimate the signal $M^*\in \mathbb{R}_+^{m_1\times m_2}$ from measurements $y\in \mathbb{Z}_+^N$.

\subsection{Regularized Maximum-Likelihood Estimator}

We propose a regularized maximum-likelihood estimator. The probability density function of $y$ is given by
\ben
p(y|\mathcal{A} M^*) =  \prod_{j=1}^N \frac{[\mathcal{A} M^*]_j^{y_j}}{y_j!} e^{-[\mathcal{A} M^*]_j}.
\een
The corresponding regularization likelihood function is given by the following optimization problem
\begin{align}
\widehat{M} &\triangleq \arg \min_{M \in \Gamma} [-\log p(y|\mathcal{A} M) + \lambda \rho (M)] \nonumber \\
& = \arg \min_{M \in \Gamma} [-\sum_{j=1}^N y_j \log [\mathcal{A} M]_j + [\mathcal{A} M]_j + \lambda \rho (M)],
\label{estimator}
\end{align}
where $\rho(M) > 0$ is a regularization function and $\lambda > 0$ is the regularization parameter. Here $\Gamma$  is a countable set of feasible estimators
\ben
\Gamma \triangleq  \{M_i \in \mathbb{R}_+^{m_1\times m_2}: \|M_i\|_{1, 1} = I, i = 1, 2, \ldots\},
\een
and the regularization function satisfies the Kraft inequality
\ben
\sum_{M \in \Gamma} e^{-\rho(M)} \leq 1.
\een
We can think of this formulation as a discretized feasible domain version of the general regularized maximum likelihood estimator. The regularization function assigns small value for lower rank $M$ and vice versa. Using Kraft-compliant regularization to prefix codes for estimators is a commonly used technique \cite{raginsky2010compressed}.

The performance metric we use for estimator is a normalized risk:
\ben
R(M^*, M) \triangleq \frac{1}{I^2} \|M^*-M\|_F^2,
\een
where $\|X\|_F^2$ is the Frobenius norm of a matrix $X$, which is defined as
$\|X\|_F^2 = \sum_i \sum_j |[X]_{ij}|^2$.

\subsection{Sensing operator}
We adopt the following linear sensing operator $\mathcal{A}$ by using the sensing matrices suggested by [6] with the form of
\ben
[A_i]_{jk} = \left\{\begin{array}{ll}
0, & \hbox{with probability } p;\\
1/N, & \hbox{with probability} 1-p.
\end{array}\right.
\een

It can be verified that this operator $\mathcal{A}$ satisfies the requirements in the previous subsection. In particular,

\noindent 1. all entries of $A_i$ take values of 0 or $1/N$;

\noindent 2. $\cA$ satisfies flux preserving. Since all entries of $A_i$ are less than $1/N$, for  $[M]_{ij}\geq 0$ for all $i,j$,
\[
\|\cA M\|_{1} = \sum_{i=1}^N \sum_{j=1}^{m_1} \sum_{k=1}^{m_2} [A_i]_{jk} M_{jk} \leq  \sum_{j=1}^{m_1} \sum_{k=1}^{m_2} M_{jk} = I.
\]
\noindent where $\|x\|_1 = \sum_i |x_i|$ denotes the $\ell_1$ norm of a vector $x$.

\noindent 3. with probability at least $1-Np^{m_1 + m_2}$, every $A_i$ has at least one non-zero entry. It follows that for $M$ such that $[M]_{ij} \geq c$, for all $i$, $j$, we have that
\[
[\cA M]_i = \sum_{j=1}^{m_1} \sum_{k=1}^{m_2} [A_i]_{jk} M_{jk} \geq c\sum_{j=1}^{m_1} \sum_{k=1}^{m_2} [A_i]_{jk}  \geq c/N.
\]
%
This operator also satisfies the restrictive isometry property.

\section{PMLSV algorithm}

In this section, we introduce a Poisson noise Maximal Likelihood Singular Value thresholding (PMLSV) algorithm for solving the regularized maximum likelihood problem formulated in (\ref{estimator}).

Nuclear norm of some matrix is defined as sum of singular values of the matrix and it is proven to be a very useful norm when solving low-rank matrix recovery problem because its close connection with rank of matrix \cite{candes2009exact} and its convexity. Therefore, it is reasonable for us to use the nuclear norm of $M$, denoted as $\|M\|_*$, for $\rho(M)$.
Therefore, we recover the low-rank matrix by solving an optimization problem
\begin{equation}
\min_{M \in \Gamma_0}  f(M) + \lambda \|M\|_{*}
\label{originaloptimizationproblem}
\end{equation}
where $f(M) = -\log p(y|\cA M)$, and we also relax the feasible domain $\Gamma$ to $\Gamma_0$
\begin{equation}
\Gamma_0 \triangleq  \{M \in \mathbb{R}_+^{m_1\times m_2}: \|M\|_{1, 1} = I \}.
\end{equation}
To solve (\ref{originaloptimizationproblem}), we may use the interior-point method since nuclear norm minimization problem with convex feasible domain can be reformulated as a Semidefinite program (SDP). However, the large number of dummy variables makes this approach less preferable as an computationally efficient algorithm for large problem. Hence, we seek an alternative approximate algorithm other than solving the SDP.

To derive the approximate algorithm, we first expand the likelihood function of our cost function by Taylor expansion and only keep up to second term as approximation. Under such approximation, (\ref{originaloptimizationproblem}) becomes
\begin{equation}
    M_k = \arg\min_M \left[Q_{t_k}(M,M_{k-1}) + \lambda\|M\|_{*}\right],
\label{ouroptimizationproblem}
\end{equation}
with
\begin{align}
    Q_{t_k}(M,M_{k-1}) &:= f(M_{k-1}) + \langle M-M_{k-1},\nabla f(M_{k-1}) \rangle \nonumber \\
    &+ \frac{t_k}{2}\|M-M_{k-1}\|_F^{2}, \label{new}
\end{align}
where $t_k$ is the step size at $k$th iteration. Note that (\ref{new}) is an optimization problem with a form similar to that studied in \cite{ji2009accelerated} and the optimizer can be derived analytically as follows.
By dropping and introducing terms independent on $M$ whenever needed (denoted by "$\propto$"), we can rewrite $Q_{t_k}(M,M_{k-1})$ as:
\begin{equation}
\begin{split}
&Q_{t_k}(M,M_{k-1}) \\
&\propto \langle M-M_{k-1},\nabla f(M_{k-1}) \rangle + \frac{t_k}{2}\|M-M_{k-1}\|_F^{2} \\
&\propto \langle M-M_{k-1},\nabla f(M_{k-1}) \rangle + \frac{t_k}{2}\|M-M_{k-1}\|_F^{2} \\
& ~~~+ \frac{t_k}{2}\| \frac{1}{t_k} \nabla f(M_{k-1}) \|_F^2\\
&\propto \frac{t_k}{2} \left\langle M-M_{k-1}+\frac{1}{t_k}\nabla f(M_{k-1}),M-M_{k-1}+\frac{1}{t_k}\nabla f(M_{k-1}) \right\rangle \\
& = \frac{t_k}{2} \left\| M-\left(M_{k-1}-\frac{1}{t_k}\nabla f(M_{k-1})\right) \right\|_F^2
\end{split}
\label{temp1}
\end{equation}
Substituting (\ref{temp1}) into (\ref{ouroptimizationproblem}) and scale the cost function by $1/t_k$, we have:
\begin{align}
    &M_k = \nonumber\\ &\arg\min_M \left[\frac{1}{2} \left\| M- \left( M_{k-1} - \frac{1}{t_k}\nabla f(M_{k-1}) \right) \right\|_{F}^{2} + \frac{\lambda}{t_k}\|M\|_{*}\right].
\label{ourfinalproblem}
\end{align}
The solution to (\ref{ourfinalproblem}) is given by a form of Singular Value Thresholding (SVT) \cite{cai2010singular}. 
Consider the following problem
\begin{equation}
\min_{Y \in \mathbb{R}^{n_1\times n_2}} \left\{ \frac{1}{2}\|Y-X\|_{F}^2+\tau\|Y\|_{*} \right\},
\label{singularvalueproblem}
\end{equation}
where $X \in \mathbb{R}^{n_1\times n_2}$ is given and $\tau$ is the regularization parameter.
%
For a matrix $X \in \mathbb{R}^{n_1\times n_2}$ with rank $r$, let its singular value decomposition be $X=U\Sigma V^T$, where $U \in \mathbb{R}^{n_1\times r}$, $V \in \mathbb{R}^{n_2\times r}$, $\Sigma=\diag(\{\sigma_i\},i=1,2,...,r)$, and $\sigma_i$ is a singular value of the matrix $X$.
For each $\tau\ \geq0$, define the {singular value thresholding operator} as :
\begin{equation}
    D_{\tau}(X) \triangleq U D_{\tau}(\Sigma)V^T,
\end{equation}
where the $D_{\tau}(\Sigma) \triangleq \diag({(\sigma_i-\tau)_{+}})$, and $(x)_{+} = \max (x,0)$.
The solution to (\ref{singularvalueproblem}) is given by singular value thresholding according to the following theorem (Theorem 2.1 in \cite{cai2010singular})\begin{thm}
For each $\tau\ \geq0$, and $X\in\mathbb{R}^{n_1\times n_2}$:
\begin{equation}
    D_{\tau}(X) = \arg \min_{Y \in \mathbb{R}^{n_1\times n_2}} \left\{ \frac{1}{2}\|Y-X\|_{F}^2+\tau\|Y\|_{*} \right\}.
\end{equation}
\label{thm_cai}
\end{thm}
Theorem \ref{thm_cai} indicates that the solution to (\ref{ourfinalproblem}) is given by
\begin{equation}
    M_k = D_{\lambda/t_k} \left( M_{k-1} - \frac{1}{t_k}\nabla f(M_{k-1}) \right).
\label{Watkthiteration}
\end{equation}
The remaining question then becomes how to deal with the feasible set $\Gamma_0$. Note that (\ref{ourfinalproblem}) is a strongly convex problem, so it is reasonable to project $M_k$ onto the convex set $\Gamma_0$ at the $k$th iteration. For a matrix $M$, define
\begin{equation}
\mathcal{P}(M)=\frac{I}{\|M\|_{1,1}} M
\end{equation}
as the projection of $M$ on to the convex set $\Gamma_0$. At the $k$th iteration, we replace $M_k$ obtained from (\ref{Watkthiteration}) by $\mathcal{P}(M_k)$. Note that at the $k$th iteration we do not force $M_k$ to be a non-negative matrix and the following initialization explains the reason.

Intuitively, the initialization we choose should be as close as possible to the matrix with maximal likelihood. In other words, we would initialize with a matrix that minimizes $f(M)$ in (\ref{originaloptimizationproblem}). For this consideration, we initialize by $M_0 = \mathcal{P}(\sum_{i=1}^n y_i A_i)$ (similar to the initialization for alternating minimization in \cite{ji2009accelerated}). However, a difference from \cite{jain2013low} is that rather than taking the top $k$ singular value, we keep all singular values to preserve information that may be needed for future iterations before truncating them prematurely.
%
%
In our algorithm, all singular values of $M_k$ decreases as $k$ increases.
With such an initialization,  the magnitude of the gradient $\nabla [-\log p(y|\cA M_{k})]$ is typically small at each iteration.  Hence, we can ensure each $M_k$ to be nonnegative by choosing a sufficiently small step size at the $k$th iteration.
The algorithm is summarized in Algorithm \ref{alg:main}.

\begin{algorithm}[h!]
  \caption{PMLSV}
  \begin{algorithmic}[1]
    %
    \STATE Initialize: $M_0 = \mathcal{P}(\sum_{i=1}^n y_i A_i)$, parameter $\gamma$, step size $L$
    \FOR{$k = 1, 2, \ldots NOI $}
    \STATE $\mathcal{G}(M_{k-1}) := \nabla [-\log p(y|\cA M_{k-1})]$
    \STATE $C := M_{k-1} - \frac{1}{L}\mathcal{G}(M_{k-1})$
    \STATE singular value decomposition: $C := UDV^T$
    \STATE $D_{\rm new} := \diag((\diag(D)-\frac{\lambda}{L})_{+})$
    \STATE $W_k := \mathcal{P}(UD_{\rm new}V^T)$.
    %
    \STATE If $F(M_{k})<F(M_{k-1})$, then $k=k+1$; else $L=\gamma L$, go to 6.
    \STATE If $|F(M_{k}) - F(M_{k-1})| < 0.5/NOI$, then $k=k-1$, exit;
    \ENDFOR
  \end{algorithmic}
  \label{alg:main}
\end{algorithm}

Details of Algorithm \ref{alg:main} are as follows. Here $L$ is the step size, $\gamma > 1$ changes the step size to ensure the cost function to decrease at each iteration, and $NOI$ is the maximum number of iterations. Steps $3$-$7$ generate solution to (\ref{ourfinalproblem}) at the $k$th iteration. Step $8$ examines if the cost function is reduced in the iteration. If the cost function does not decrease,
we update the step size by multiplying $\gamma$ in order to change the singular value more conservatively. In Step $9$, if the absolute difference of cost function between consecutive two iterations is less than $0.5/NOI$, then we stop the algorithm. The choice of $NOI$ is user specified: a larger $NOI$ leads to more accurate solution, and a small $NOI$ obtains the solution quickly at the cost of accuracy.

\section{Examples}

We use the image of solar flare as example (see \cite{XieHuang2013} for detailed explanation of the data). We break the image into 8 by 8 patches and vectorize each patch to be a column of a new matrix. This new matrix formed by vectorized patches can be well approximated using a low-rank matrix, as demonstrated in Fig. 1. The intensity of the image is $I=2.37\times 10^7$. To change SNR of the image, we scale the image intensity by $\alpha\geq 1$. The parameters for the PMLSV algorithm are $L = 10^{-5}$, $\gamma=1.1$, and $NOI=2500$.

\begin{figure}[h]
\begin{tabular} {cc}
\includegraphics[width = 0.3\linewidth]{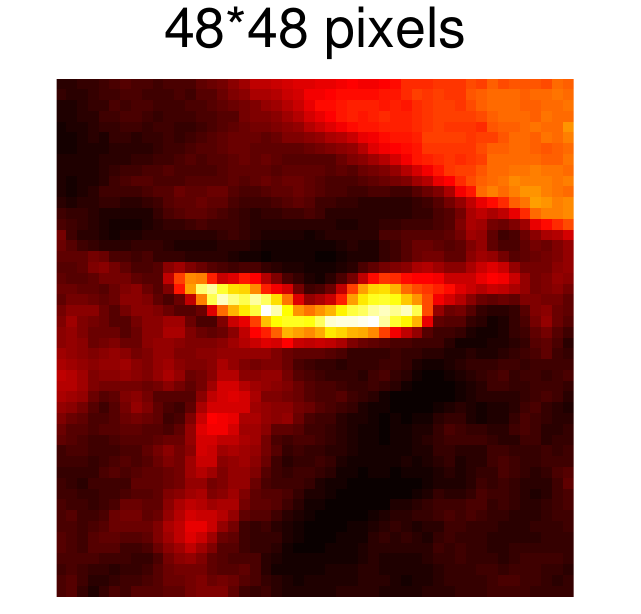} & \includegraphics[width = 0.3\linewidth]{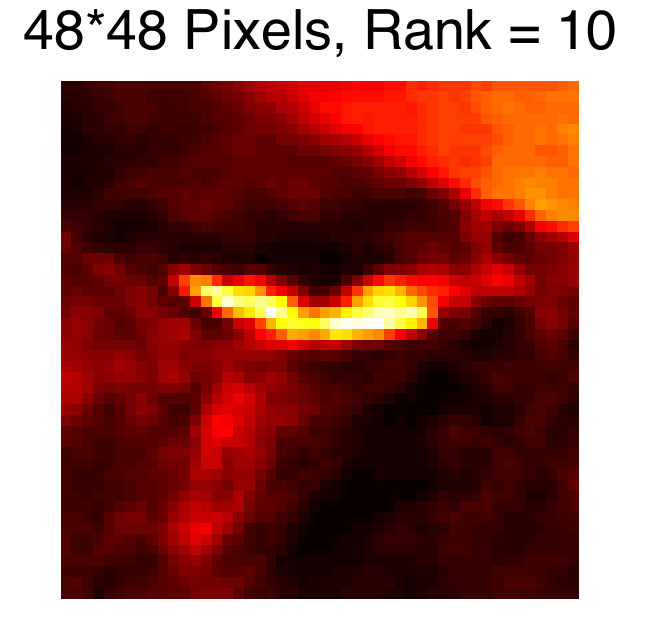} \\
(a) original solar flare image & (b) solar flare image with rank 10
\end{tabular}
\caption{Original and low-rank solar flare image.}
\end{figure}

First, we run the PMLSV algorithm and solve the SDP using CVX\footnote{http://cvxr.com/cvx/}  with various number of measurements, respectively,  when fixing $\alpha=4$ and $\lambda=0.002$. In Fig. 2, the blue line represents the risk by running PMLSV algorithm and red line represents the risk by running SDP given the same observations, respectively. Fig. 2 demonstrates that more measurements lead to more smaller risk, as expected. Also, since it is an approximate algorithm, PMLSV algorithm is less accurate than SDP; however, the maximal increase in risk of PMLSV algorithm relative to that of SDP is 4.89\%.  PMLSV is much faster: as shown in Table 1 which is the CPU running time of solving SDP by CVX and our PMLSV algorithm. 

\begin{figure}[h]
\begin{center}
\includegraphics[width = 1.0\linewidth]{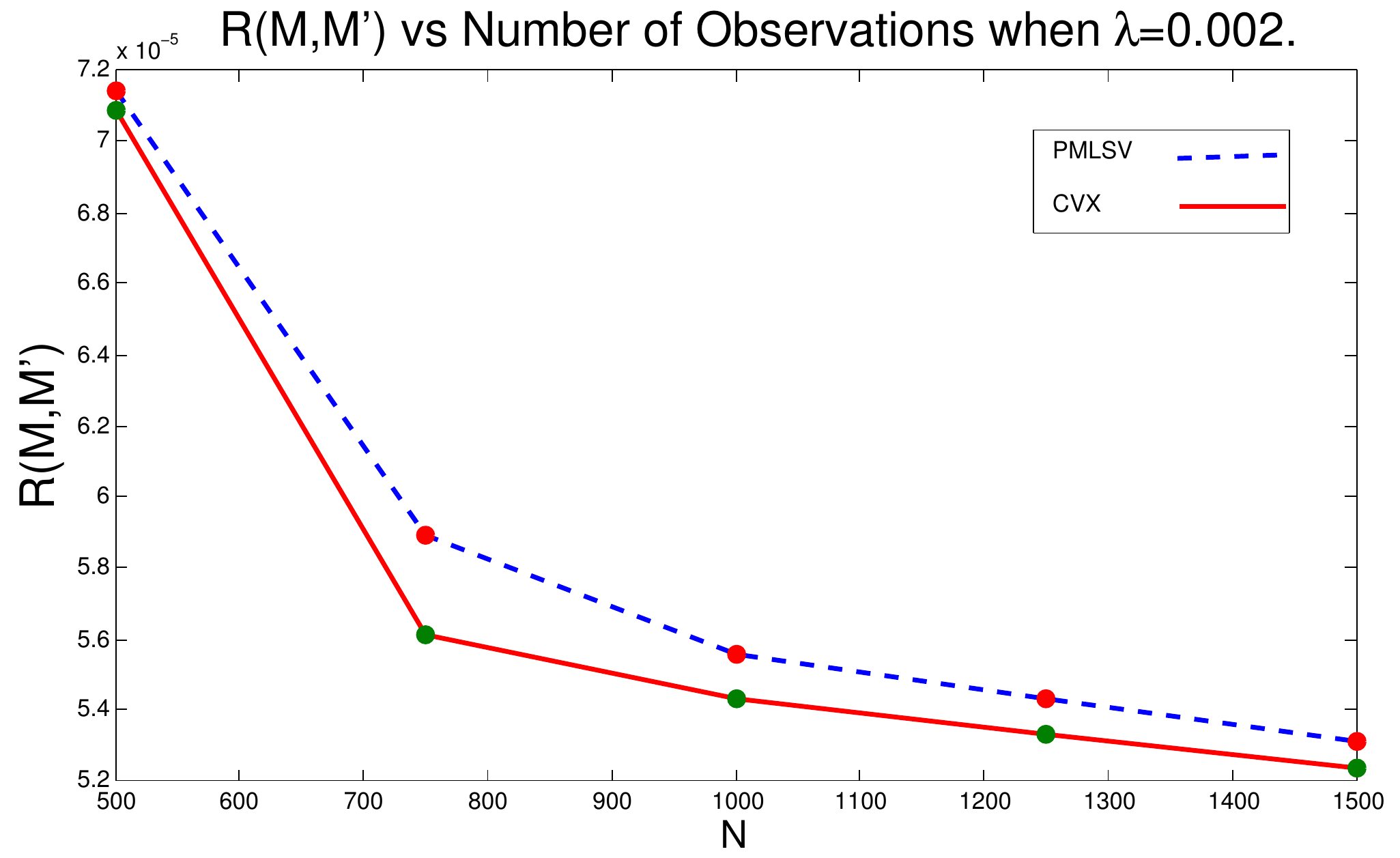}
\caption{Risk vs the number of measurements when $\alpha=4$ and $\lambda=0.002$. Points from left to right correspond to risk when $N=500,750,1000,1250, \rm and 1500$, respectively, and solved by CVX and PMLSV.}
\end{center}
\end{figure}

\begin{table}[h]
\center
\caption{CPU time (in seconds) of solving SDP by using CVX and our PMLSV algorithm when fixing $\alpha=4$ and $\lambda=0.002$ with 500, 750, 1000, 1250 and 1500 measurements, respectively.}
\begin{tabular}{|l|c|c|c|c|c|}
  \hline
  $N$ & 500 & 750 & 1000 & 1250 & 1500 \\
  \hline
  SDP & 725s & 1146s & 1510s & 2059s & 2769s \\
  \hline
  PMLSV & 172s & 232s & 378s & 490s & 642s \\
  \hline
\end{tabular}
\end{table}

Second, we run our algorithms with different $\alpha$ when fixing $N=1000$ and $\lambda=0.002$. The results are shown in Fig. 3. The larger $\alpha$ (hence the higher the SNR) we have, the lower the risk as demonstrated in Fig. 4.

\begin{figure}[h]
\begin{center}
\includegraphics[width = 0.8\linewidth]{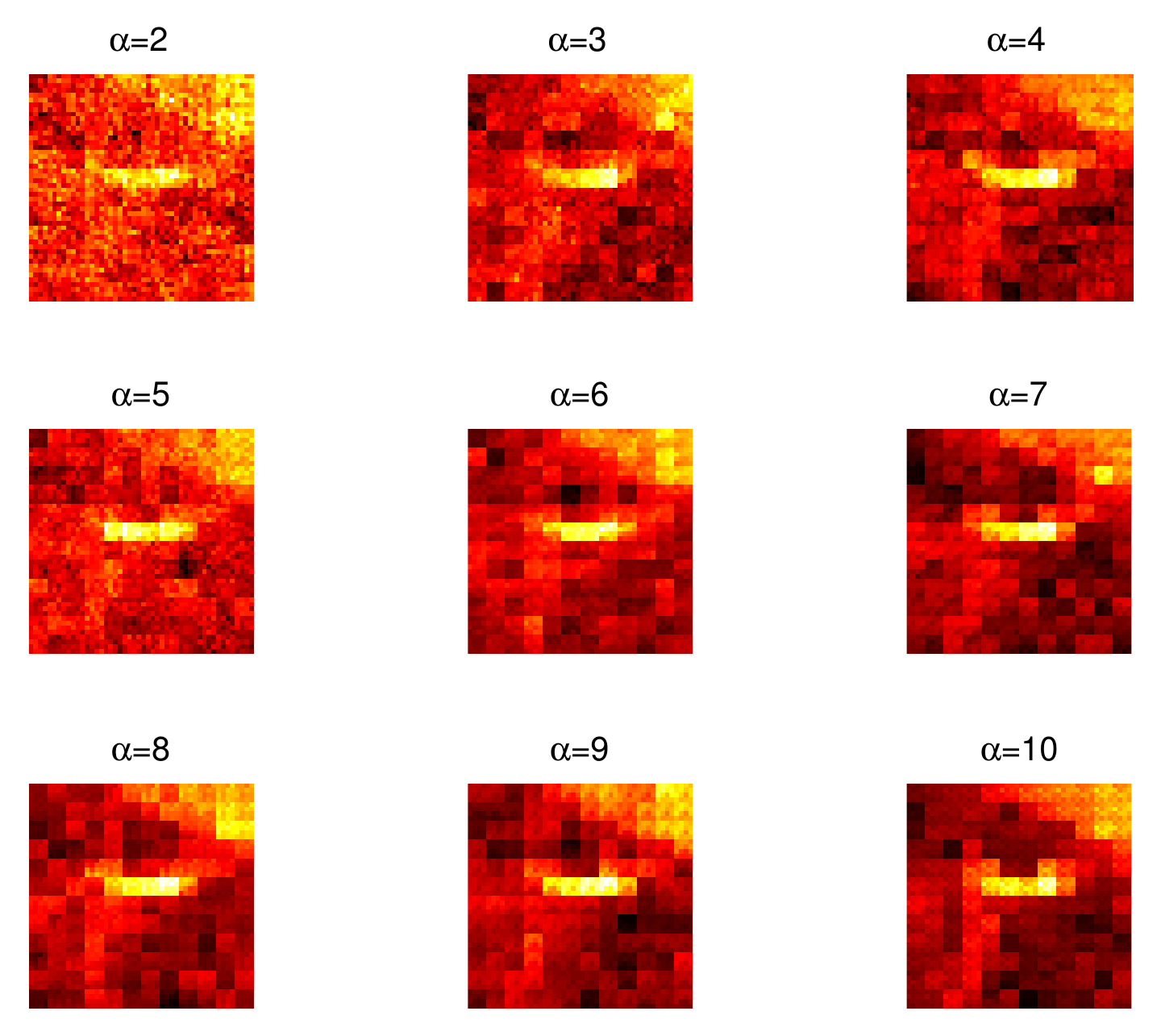}
\caption{Recovery results when fixing $N=$1000, $\lambda=0.002$ with different value of $\alpha$.}
\end{center}
\end{figure}

\begin{figure}[h]
\begin{center}
\includegraphics[width = 1.0\linewidth]{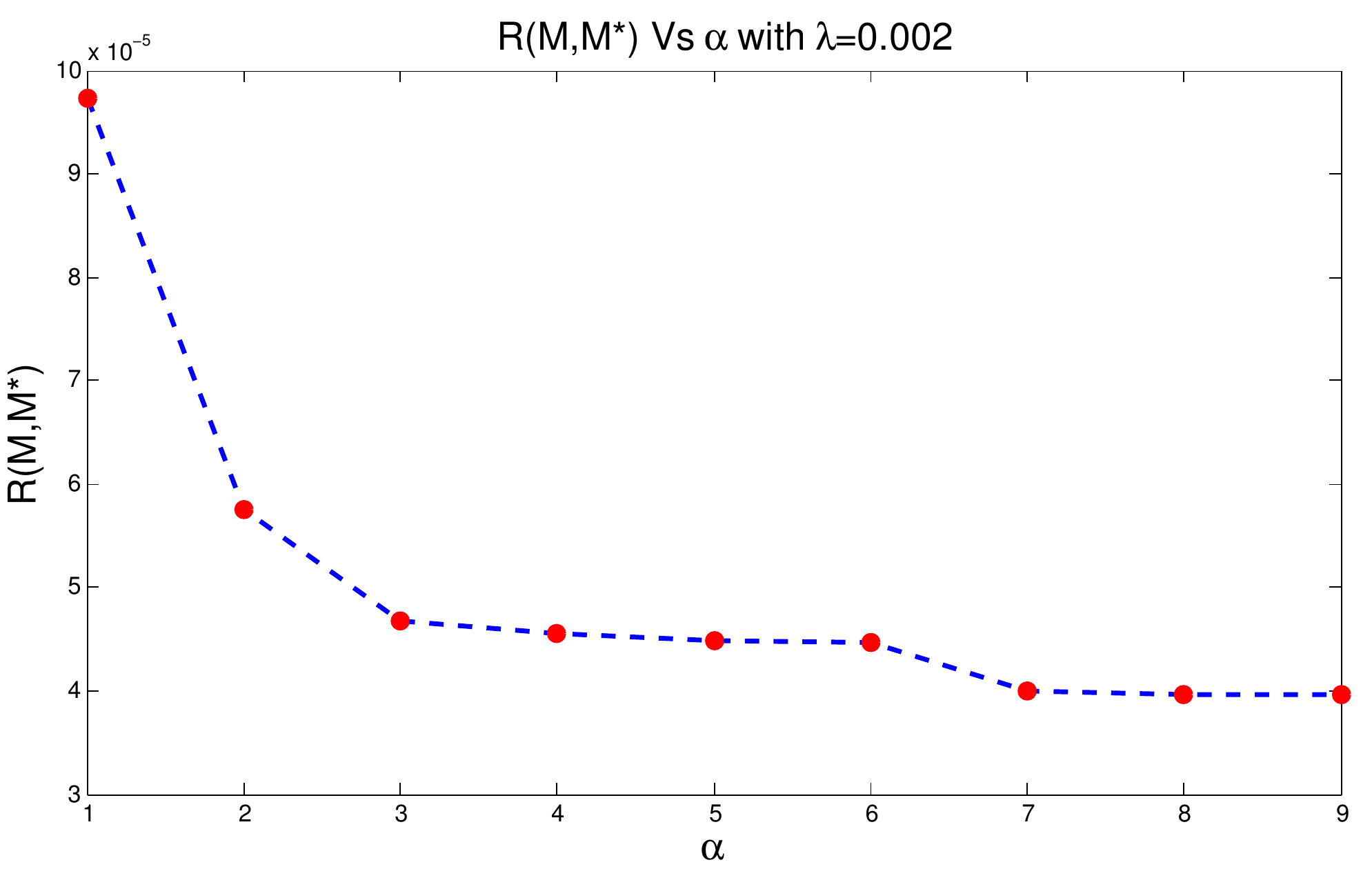}
\caption{Risk vs $\alpha$ when fixing $N=1000$, $\lambda=0.002$. Points from left to right means the risk with $\alpha=1$ to $\alpha=9$.}
\end{center}
\end{figure}

Third, we run our algorithm with different values of $\lambda$ when fixing $N=1000$ and $\alpha=4$. The results are shown in Fig. 5. From Fig. 6, we can see that there is an optimal value for $\lambda$ which leads to the smallest risk.

\section{Conclusion and future work}

We have presented a new algorithm for low-rank matrix recovery with linear measurements contaminated with Poisson noise: the Poisson noise Maximal Likelihood Singular Value Thresholding (PMLSV) algorithm, based on solving a regularized maximum likelihood problem with nuclear norm as the reguarlizer. We demonstrate its accuracy and efficiency compared with the semi-definite program (SDP) and tested on real data examples of solar flare images. Future work include analyzing the convergence property of the algorithm, and extension to the related matrix completion problem with Poisson noise.

\begin{figure}[h!]
\begin{center}
\includegraphics[width = 0.8\linewidth]{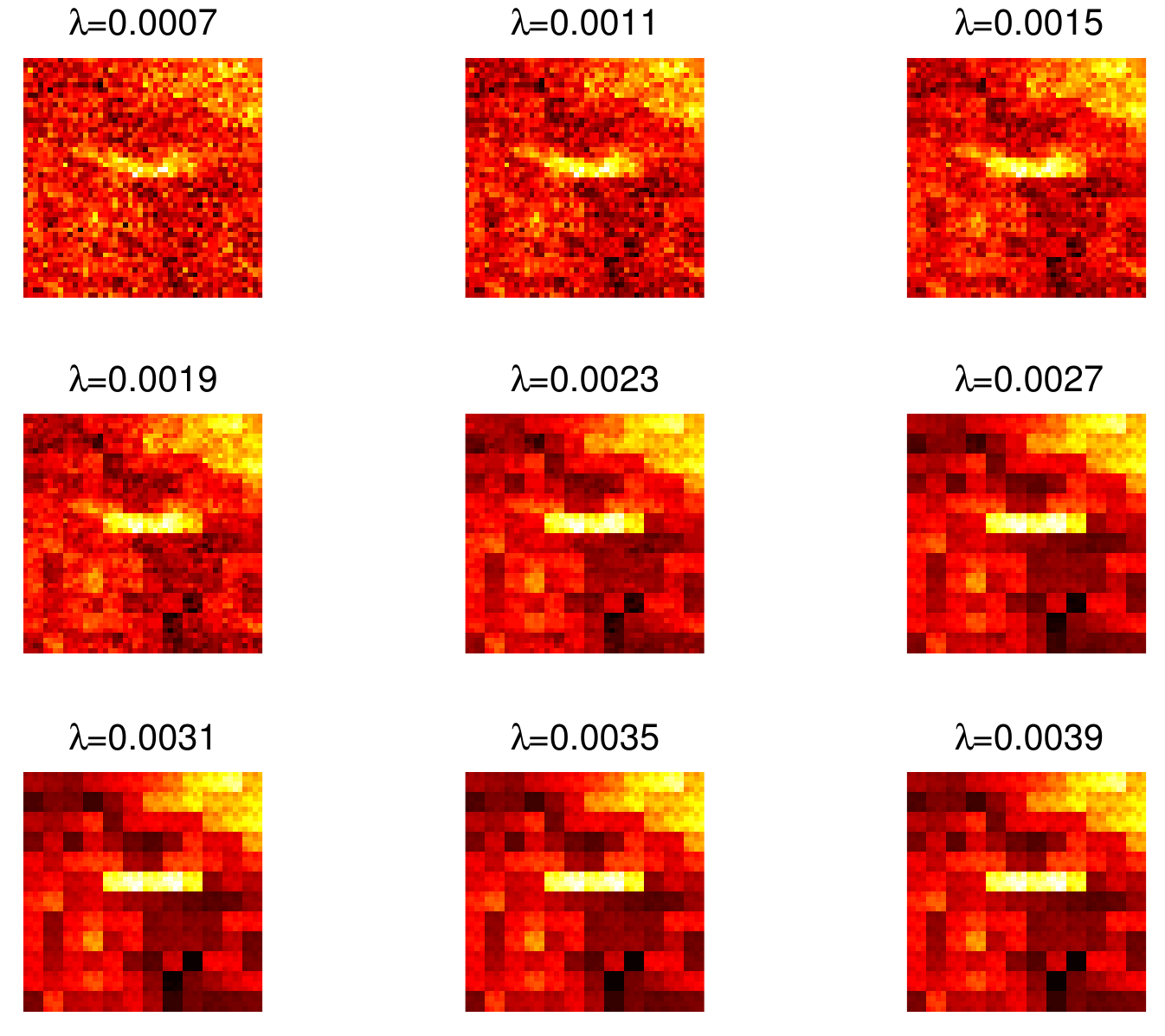}
\caption{Recovery results with different value of $\lambda$ when fixing $N=1000$ and $\alpha=4$.}
\end{center}
\end{figure}

\begin{figure}[h!]
\begin{center}
\includegraphics[width = 1.0\linewidth]{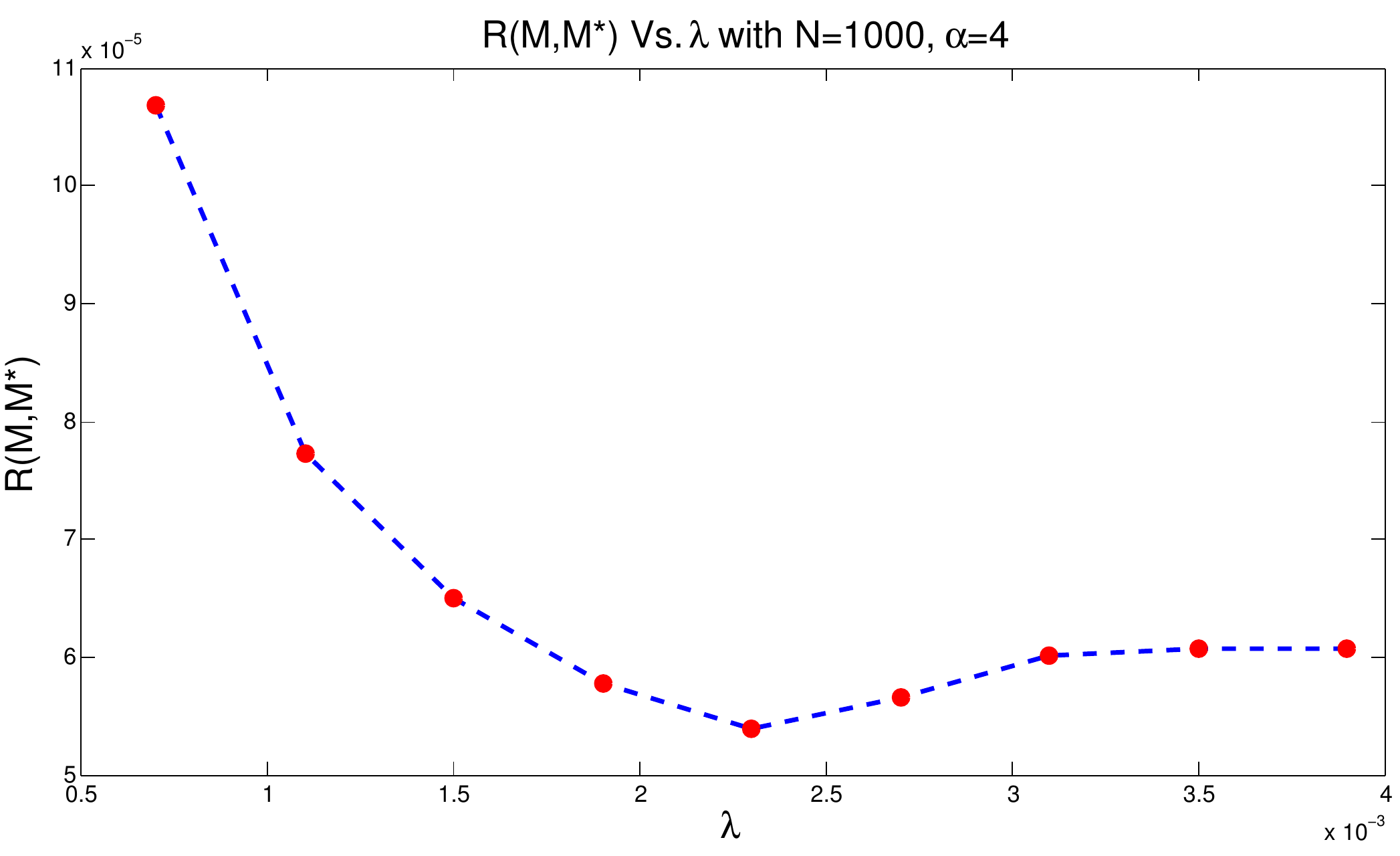}
\caption{Risk vs $\lambda$ when $N=1000$, $\alpha=4$. Points from left to right means the risk when $\lambda$=0.0007 to $\lambda$=0.0039 with step size 0.0004.}
\end{center}
\end{figure}

\bibliography{PoissonMC}

\end{document}